\documentclass{article} 
\usepackage{iclr2023_conference,times}

\usepackage{amsmath,amsfonts,bm}

\def\eqref#1{equation~\ref{#1}}

\def\1{\bm{1}}

\def\vc{{\bm{c}}}

\def\vt{{\bm{t}}}

\def\vw{{\bm{w}}}
\def\vx{{\bm{x}}}
\def\vy{{\bm{y}}}
\def\vz{{\bm{z}}}

\def\mA{{\mathbf{A}}}

\def\mT{{\mathbf{T}}}
\def\mU{{\bm{U}}}
\def\mV{{\mathbf{V}}}
\def\mW{{\mathbf{W}}}
\def\mY{{\mathbf{Y}}}
\def\mZ{{\bm{Z}}}

\DeclareMathAlphabet{\mathsfit}{\encodingdefault}{\sfdefault}{m}{sl}
\SetMathAlphabet{\mathsfit}{bold}{\encodingdefault}{\sfdefault}{bx}{n}

\usepackage[colorlinks=true,citecolor=blue,]{hyperref}
\usepackage{url}

\usepackage{blindtext}
\usepackage{multirow}
\usepackage{graphicx}
\usepackage{bm}
\usepackage{bbm} 
\usepackage{comment}
\usepackage{booktabs}
\usepackage{algorithm}
\usepackage{algorithmic}
\usepackage{array} 
\usepackage[safe]{tipa} 
\usepackage{color, colortbl}
\usepackage[misc,geometry]{ifsym} 
\usepackage{pifont,makecell,subcaption}
\usepackage{wrapfig}

\providecommand{\eg}{\textit{e.g.}\@\xspace}
\providecommand{\ie}{\textit{i.e.}\@\xspace}

\newcommand{\snowflake}{\includegraphics[width=10px]{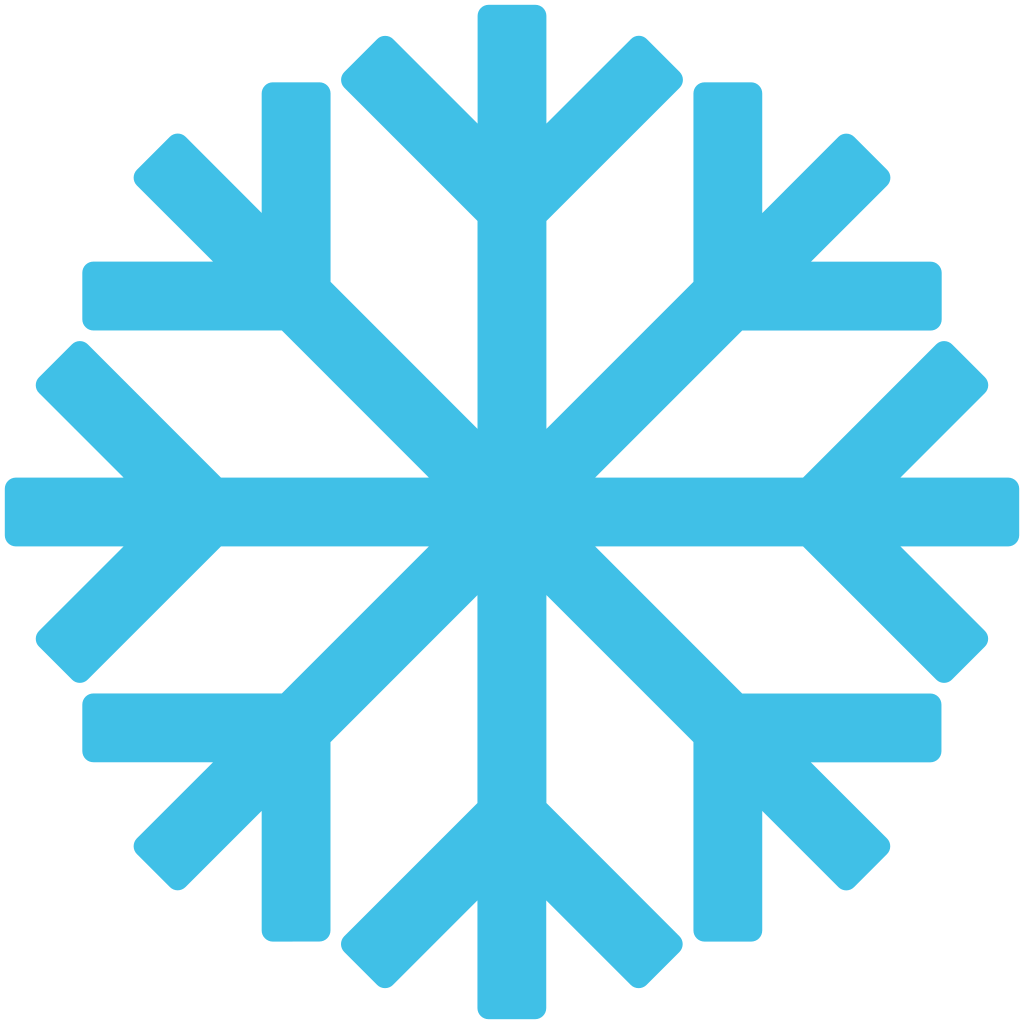}}
\newcommand{\fire}{\includegraphics[width=10px]{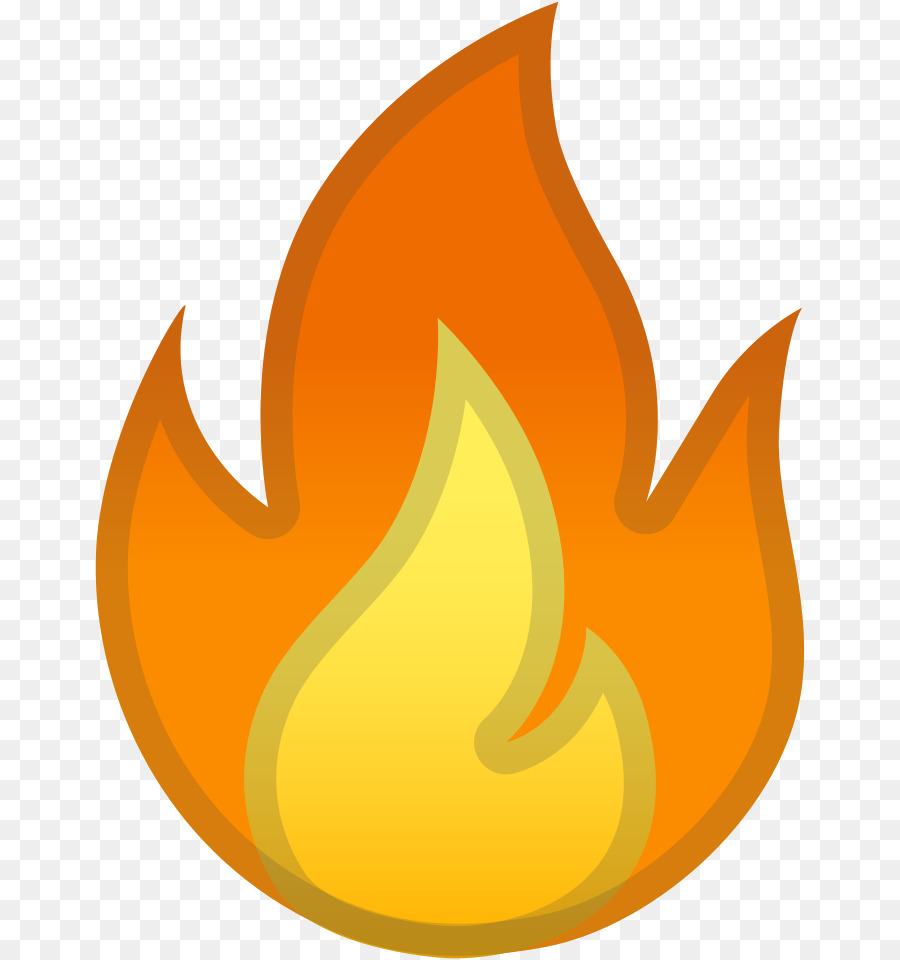}}

\makeatletter
\renewcommand\paragraph{
  \@startsection{paragraph} 
  {4} 
  {\z@} 
  {.5em \@plus1ex \@minus.2ex} 
  {-1.5em} 
  {\normalfont\normalsize\bfseries} 
}
\makeatother

\makeatletter
\def\@fnsymbol#1{\ensuremath{\ifcase#1\or \textsuperscript{~\Letter}\or \ddagger\or
   \mathsection\or \mathparagraph\or \|\or **\or \dagger\dagger
   \or \ddagger\ddagger \else\@ctrerr\fi}}
\makeatother

\newcommand{\tableCellHeight}{1}
\newcommand{\tabstyle}[1]{
  \setlength{\tabcolsep}{#1}
  \renewcommand{\arraystretch}{\tableCellHeight}
  \centering
  \small
}

\newcommand{\rotbox}[1]{\rotatebox{90}{#1}}
\newcommand{\rowNumber}[1]{\textcolor{Cerulean}{#1}}

\definecolor{ForestGreen}{RGB}{34,139,34}
\definecolor{NavyBlue}{rgb}{0.0, 0.0, 0.5}
\definecolor{rred}{RGB}{245, 152, 153}
\definecolor{oorange}{RGB}{253, 205, 154}
\definecolor{blueline}{RGB}{31, 119, 180}
\definecolor{Cerulean}{rgb}{0.0, 0.48, 0.65}

\definecolor{tabhighlight}{HTML}{e5e5e5}
\definecolor{citecolor}{HTML}{0071bc}
\definecolor{prompt_blue}{HTML}{1f78b4}
\definecolor{prompt_red}{HTML}{d45c43}

\newcommand{\methodname}{unified prompt\xspace}
\def\mycmd{1}

\iclrfinalcopy
\title{Unified Vision and Language Prompt Learning}

\author{
{
Yuhang Zang$^{1}$, Wei Li$^{1}$, Kaiyang Zhou$^{1}$, Chen Huang$^{2}$, Chen Change Loy$^{1}$\textsuperscript{\Letter}} 
\\
{
${^1}$S-Lab, Nanyang Technological University ~~${^2}$Apple Inc.
}
\\
{\tt\small $\{$zang0012, wei.l, kaiyang.zhou, ccloy$\}$@ntu.edu.sg ~chen-huang@apple.com}
}

\begin{document}

\maketitle

\begin{abstract}
Prompt tuning, a parameter- and data-efficient transfer learning paradigm that tunes only a small number of parameters in a model's input space, has become a trend in the vision community since the emergence of large vision-language models like CLIP. We present a systematic study on two representative prompt tuning methods, namely text prompt tuning and visual prompt tuning. A major finding is that none of the unimodal prompt tuning methods performs consistently well: text prompt tuning fails on data with high intra-class visual variances while visual prompt tuning cannot handle low inter-class variances. To combine the best from both worlds, we propose a simple approach called Unified Prompt Tuning (UPT), which essentially learns a tiny neural network to jointly optimize prompts across different modalities. Extensive experiments on over 11 vision datasets show that UPT achieves a better trade-off than the unimodal counterparts on few-shot learning benchmarks, as well as on domain generalization benchmarks. Code and models will be released at {\small\url{https://github.com/yuhangzang/UPT}} to facilitate future research.
\end{abstract}
\section{Introduction} \label{sec:introduction}

Vision-language~(VL) models (\eg, CLIP~\citep{radford2021learning} and ALIGN~\citep{jia2021scaling}) pre-trained on millions of image-text pairs have shown promising transferability on various downstream tasks such as few-shot learning~\citep{zhou2021coop,zhou2022cocoop,ju2021prompting} and open-vocabulary perception~\citep{gu2021open,zhou2022detecting,zang2022open,ghiasi2021open}.
When adapting large VL models to downstream tasks, it is often prohibitive to fine-tune the entire model directly due to their huge parameter size.
Therefore, many studies~\citep{gao2021making,li2021prefix,lester2021power,zhou2021coop,lu2022prompt,ju2021prompting,yao2021cpt,jia2022visual,bahng2022visual} explore prompt tuning, \ie, freezing the parameters of VL models and only fine-tuning extra learnable parameters, known as \textit{prompts}, to adapt VL models for downstream tasks efficiently and effectively.

\begin{figure*}[t]
\centering
\includegraphics[width=\textwidth]{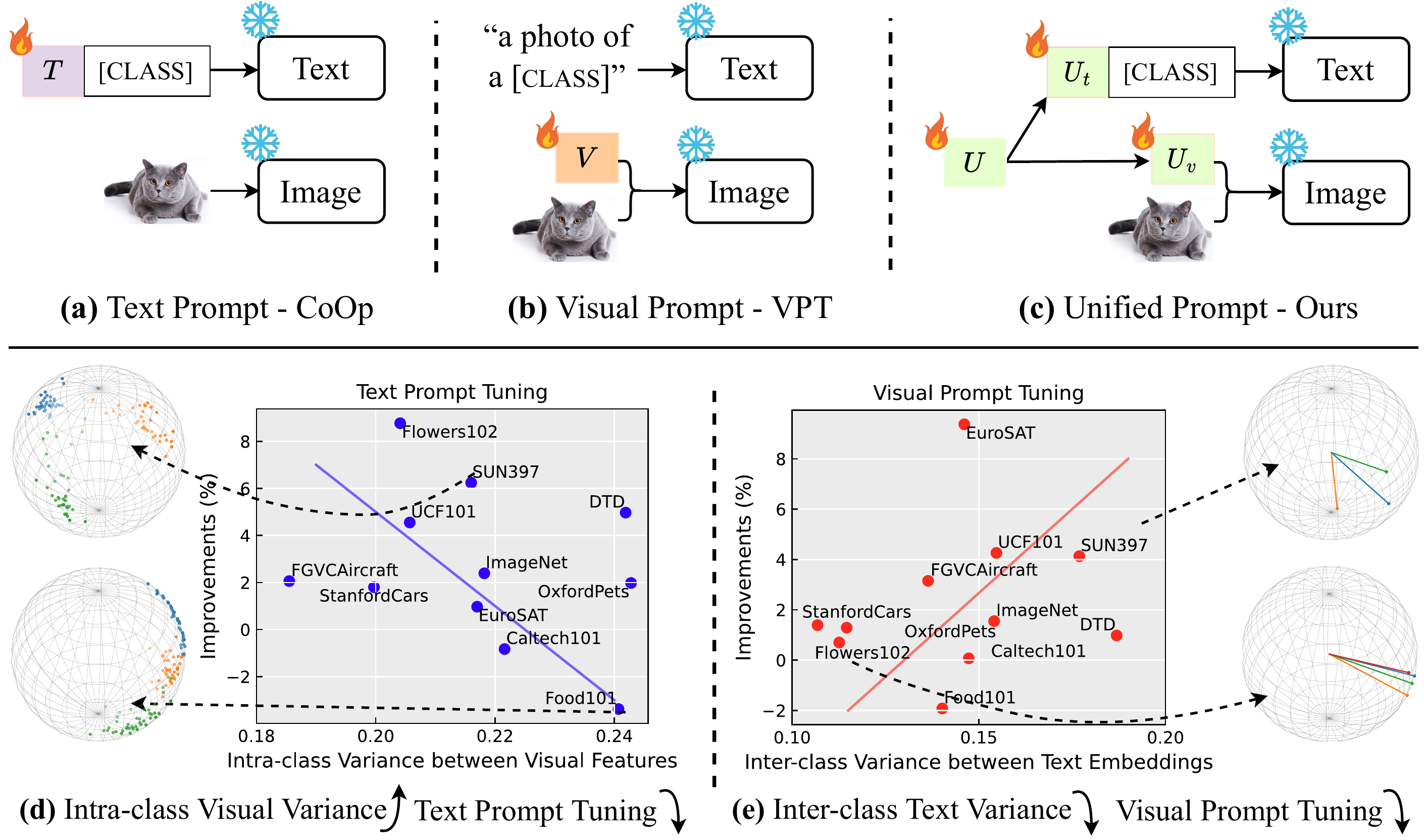}
\caption{\textbf{Top}: the architecture paradigm of \textbf{(a)} \emph{text} prompt tuning~\citep{zhou2021coop}, \textbf{(b)} \emph{visual} prompt tuning~\citep{jia2022visual} and \textbf{(c)} our multi-modal \emph{unified} prompt tuning (\fire: learnable; \snowflake: frozen parameters). \textbf{Bottom}: the performance improvements~(\%) of \emph{text} prompt tuning~\textbf{(d)} and visual prompt tuning~\textbf{(e)} compared with the zero-shot CLIP baseline. We show that the variance of \emph{visual} and \emph{text} features ($x$-axis) will affect the improvements ($y$-axis).
We project the text/visual features of the dataset (pointed by the dashed arrow) into a unit sphere to show the variance of different distributions.
Please refer to the appendix for the implementation details about how we compute the feature variance.
}
\label{fig:comparison}
\vspace{-12pt}
\end{figure*}

A typical VL model consists of two sub-networks---an image encoder and a text encoder---to extract representations for visual and text modalities.
Correspondingly, existing prompt tuning approaches can be grouped into two types: text prompt tuning and visual prompt tuning.
For text prompt tuning methods, such as CoOp~\citep{zhou2021coop}), extra text prompts, treated as learnable parameters, are applied on the \emph{text encoder} in CLIP (Fig.~\ref{fig:comparison}(a)), to mitigate the potentially sub-optimal hand-crafted text prompt templates (\eg, ``\texttt{a photo of a {[CLASS]}.}''). 
On the contrary, visual prompt tuning approaches focus on modulating the \emph{image encoder}~(Fig.~\ref{fig:comparison}(b)). For example, VPT~\citep{jia2022visual} injects learnable parameters, known as visual prompts, into multiple layers of vision Transformers. 
In general, these prompt approaches consider tuning the representations of visual and text modalities independently.

Despite significant improvements that have been achieved, we observe that existing prompt tuning approaches~\citep{zhou2021coop,jia2022visual} cannot obtain consistent performance improvements due to inherent variances in visual features and text embedding in downstream tasks.
That is to say, using the single-modal prompt may acquire good results on one task while performing poorly on other tasks.
To analyze this phenomenon, we measure the discrepancy in data distribution via the intra-class variance of \emph{visual} features and inter-class variance of \emph{text} embedding and study the correlation between data statistics and performance improvements. 
As shown in Fig.~\ref{fig:comparison}(d), when the intra-class variance of image features
are large (bottom right), we observe that CoOp struggles to learn suitable text prompts for improving the text classifier.
As for visual prompt tuning, VPT faces difficulties when inter-class variance of text embeddings are low, as shown in bottom left of Fig.~\ref{fig:comparison}(e). That is, if the text classifiers are established based on text embeddings of low separability, tuning visual prompts would lend little help to improve the final performance.
Moreover, intra-class visual variance and inter-class text variance are typically orthogonal.
Thus, we can observe that single-modal prompt tuning methods vary widely across different datasets: CoOp beats VPT with 8.1\% on Flowers102~\citep{nilsback2008automated}, while VPT outperforms CoOp by 8.4\% on EuroSAT~\citep{helber2019eurosat}. Choosing which prompt modality to tune becomes a dilemma given different downstream tasks. 
Clearly, the key is to simultaneously adapt both text and visual prompts to overcome the vast differences across different data distributions.
A straightforward solution is to introduce both \emph{text} and \emph{visual} prompts to the model and jointly optimize two modality-specific prompts together.
However, we find that such a na\"{i}ve joint training leads to sub-optimal performance due to the intrinsic discrepancy between text and image modalities. In particular, the performance is occasionally worse than tuning modality-specific prompts as shown in our experiments.

Solving the issues above requires modality-agnostic optimization to bridge the isolated prompts. To this end, we present a unified prompt tuning for both the \textit{text} and \textit{visual} modalities, dubbed as \textbf{U}nified \textbf{P}rompt \textbf{T}uning~(\textbf{UPT}), see Fig.~\ref{fig:comparison}(c).
Specifically, we start with a shared initial prompt and propose a lightweight self-attention network to generate the prompt for CLIP \textit{text} and \textit{visual} encoders.
We empirically show that such a design can preserve the benefit of individual modality.

Our contributions are summarized as follows: \textbf{1)} We provide a comprehensive analysis for existing \emph{text} or \emph{visual} prompt tuning strategies; \textbf{2)} We present a unified prompt method for VL models to tune both the \emph{visual} and \emph{text} modality representations; \textbf{3)} We conduct extensive experiments to show that unified prompt tuning is a viable strategy that outperforms previous single-modal prompt tuning methods, especially under the few-shot learning and domain generalization settings. We hope our work can motivate future research using multi-modal prompts for VL models.

\section{Methodology} 

We first introduce vision-language models focusing on CLIP~\citep{radford2021learning}, in company with text/visual prompt tuning approaches for visual recognition in Sec.~\ref{sub:preliminaries}.
We then analyze the limitations of previous single-modal prompt tuning approaches in Sec.~\ref{sec:analysis}. Finally, we present technical details of our proposed \methodname learning in Sec.~\ref{sec:unified_prompt}.

\subsection{Preliminaries} \label{sub:preliminaries}
\noindent \textbf{CLIP.} CLIP~\citep{radford2021learning} consists of two sub-networks: an image encoder $\phi$ and a text encoder $\psi$. These two encoders, respectively, map the text and image inputs into a joint hidden space $\mathbb{R}^{d}$, where the semantics of vision and language modalities are well-aligned. 
Here, $d$ refers to the final hidden dimension of the text or image encoder (\eg, $d=256$ in the ResNet~\citep{he2016deep} backbone and $d=512$ in the ViT backbone). 
Given an input image $\vx$ and a set of categories $\mY=\{\vy_1, \vy_2, ..., \vy_k\}$ (\eg, $k=1000$ for ImageNet~\citep{deng2009imagenet}), the image encoder extracts the corresponding image feature $\vz = f_\phi(\vx) \in \mathbb{R}^{d}$. While the class names in $\mY$ are first filled into a hand-crafted text prompt template \texttt{a photo of a [CLASS]} to obtain the text descriptions $\mA$, further processed by the text encoder for the text representations: $\mW = f_\psi(\mA) \in \mathbb{R}^{d \times k}$. The final prediction is computed as follows:
\begin{equation}
    \label{eq:cos}
    p(y=i \mid \vx)=\frac{\exp \left(\cos \left(\vw_{i}, \vz\right) / \tau\right)}{\sum_{j=1}^k \exp \left(\cos \left(\vw_{j}, \vz\right) / \tau\right)},
\end{equation}
where $\cos(\cdot, \cdot)$ denotes the cosine similarity and $\tau$ is a fixed temperature value (\eg, $\tau=100$).
Conceptually, such a decision process for the input image $\vx$ in Eq.~(\ref{eq:cos}) is formulated in a way that the text encoder $\psi$ takes a role of generating dynamic \textbf{classifiers} $\mW$ from open-set categories $\mY$, with the image encoder $\phi$ producing encoded visual \textbf{features} $\vz$. In practice, it is generally infeasible to fine-tune the millions of parameters (\ie, $\phi$ and $\psi$) in a VL model for transfer learning in every downstream task.

\noindent \textbf{Text Prompt Tuning.} For efficient and effective model adaptation, text prompt tuning approaches consider generating more adaptive \textbf{classifiers} without fine-tuning the text encoder $\psi$. 
For example, Context Optimization (CoOp)~\citep{zhou2021coop} introduce a set of learnable parameters $\mT \in \mathbb{R}^{d \times m}$ to replace the hand-crafted text prompt template (\texttt{a photo of a {[CLASS]}}). The word-embedding of class names in $\mY$ will concatenate with these text prompts in the following form:
\begin{equation} \label{eq:text_prompt}
    \hat{\mT} = [\vt_1, \vt_2, \ldots, \vt_m, \texttt{CLASS}].
\end{equation}
Here, the symbol $m$ denotes the prompt length. The resulting dynamic text representations are extracted by the text encoder: $\mW = f_\psi(\hat{\mT}) \in \mathbb{R}^{d \times k}$. In each downstream task, the learnable prompts $\mT$ will be optimized with each task-specific objective function, \eg, a cross-entropy classification loss $\mathcal{L}_{\text{CE}}(p, y)$ in few-shot learning. Note that both the image and text encoders ($\phi$ and $\psi$) are frozen during downstream training. As a result, updating the text prompt $\mT$ will correspondingly adjust the decision boundaries with generated classifiers $\mW$ for downstream tasks.

\noindent \textbf{Visual Prompt Tuning.} Conversely, visual prompt tuning methods focus on extracting more transferable visual \textbf{features} while keeping the visual encoder $\phi$ unchanged. Following the success of text prompt tuning approaches, recent Visual Prompt Tuning (VPT)~\citep{jia2022visual} introduces a similar prompt tuning recipe for the visual encoder $\phi$. Suppose the image encoder $\phi$ contains $L$ Vision Transformer layers, the output of $i$-th layer, $l_{i}$, where $i=1,2, \ldots, L$, is given by:
\begin{equation}\label{eq:vit}
    [\vc^{i+1}, \vz^{i+1}_{1}, \ldots, \vz^{i+1}_{s}]=l_{i}\left(\left[\vc^{i}, \vz^{i}_{1}, \ldots, \vz^{i}_{s} \right]\right),
\end{equation}
where $\vc \in \mathbb{R}^{d}$ denotes the classification token (\texttt{[CLS]}), and $\mZ = [\vz_1, \vz_2, \ldots, \vz_s] \in \mathbb{R}^{d \times s}$ denotes the input image patch tokens with length $s$. For the $i$-th encoder layer, a set of learnable visual prompts $\mV^i \in \mathbb{R}^{d \times n}$ are inserted and computed as follows:
\begin{equation}\label{eq:visual_prompt}
    [\vc^{i+1}, \ldots, \mZ^{i+1}]=l_{i}\left(\left[\vc^{i}, \mV^{i}, \mZ^{i}\right]\right),
\end{equation}
where $n$ stands for the length of visual prompts. Two VPT variants are proposed: VPT-shallow and VPT-deep. For VPT-shallow, the visual prompts are only inserted into the first Transformer layer ($i=1$). Whereas for VPT-deep, visual prompts are introduced at every layer.
The learnable visual prompts are data-independent, which once learned, can modulate the visual features $\vz$ of input images for better downstream transfer learning.

\subsection{Analysis} \label{sec:analysis}
\begin{figure*}[t]
\centering
\includegraphics[width=.98\textwidth]{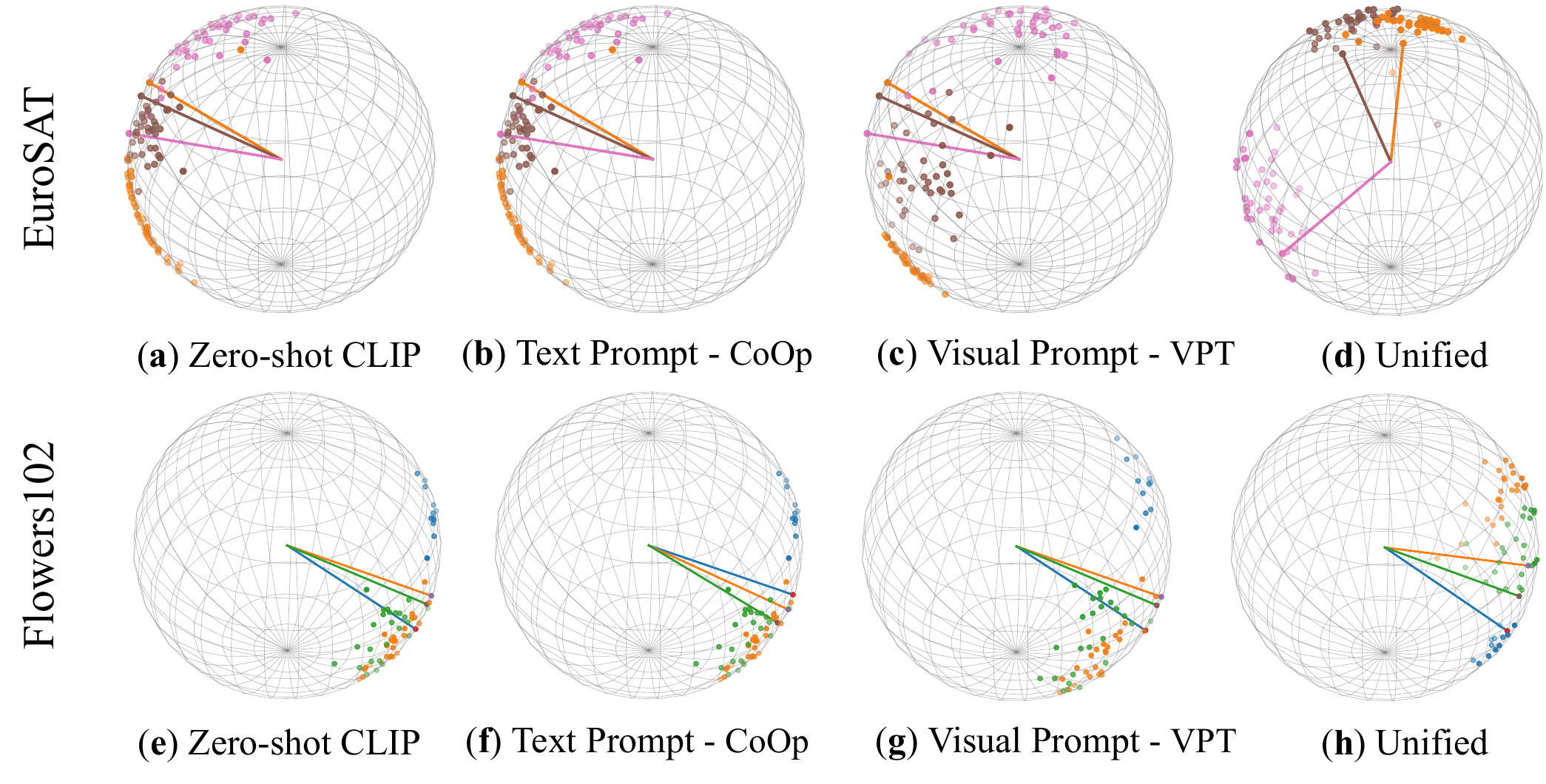}
\caption{Visualization of input features $\vz$ (projected points) and text classifier $\mW$ (projected lines) on EuroSAT and Flowers102. 
}
\label{fig:motivation}
\end{figure*}
We conduct a series of probing studies to analyze the characteristics of text/visual prompt tuning. First, when adapting the CLIP model with two representative text and visual prompt tuning approaches (CoOp~\citep{zhou2021coop} and VPT~\citep{jia2022visual}), we measure the variance of both visual features $\vz$ and text embeddings $\mW$ (\ie, classifiers) for all 11 downstream vision datasets (see Appendix~\ref{sec:app_var} for detailed implementations). For text prompt tuning, as shown in Fig.~\ref{fig:comparison}(d), we observe that CoOp performs well on datasets with low intra-class variance between visual features, such as Flowers102, but fails on Food101 dataset with high intra-class feature variance. 
As for visual prompt tuning, VPT succeeds in improving performance on SUN397 dataset with large inter-class text embeddings, while being less effective on Food101 and Flowers102 with relatively smaller inter-class text embedding variance. The performance improvements of text/visual prompt tuning are highly correlated with the variance of visual features $\vz$ or text embeddings $\mW$ in downstream datasets.

In order to understand this phenomenon, we select two downstream vision datasets (Flowers102~\citep{nilsback2008automated}, EuroSAT~\citep{helber2019eurosat}) for further analysis. During downstream training, we project both visual features $\vz$ and text embeddings $\mW$ (\ie, classifiers) into joint sphere space $\mathbb{R}^{3}$ for better visualization. As we illustrated in 
Fig.~\ref{fig:motivation}, we can observe that: \textbf{1}) For the EuroSAT dataset with high intra-class visual feature variance, text prompts in CoOp fails to adapt the text classifiers $\mW$. Clearly, the text classifiers in Fig.~\ref{fig:motivation}(\textbf{b}) are almost unchanged compared with zero-shot CLIP baseline (Fig.~\ref{fig:motivation}(\textbf{a})). \textbf{2}) For the Flowers102 dataset with low inter-class text embedding variance, visual prompts in VPT are not effective in modulating the visual features $\vz$ (Fig.~\ref{fig:motivation}(\textbf{g})), thus cannot obtain considerable performance gain.

In conclusion, the single-modal prompt tuning approaches (CoOp and VPT), face the dilemma that consistent improvements over Zero-shot CLIP are hard to achieve due to inherent variances of visual features and text embedding in downstream tasks.
Our observation motivates us to present a unified prompt tuning method that tunes the $\vz$ and $\mW$ at the same time.

\subsection{Unified Prompt Tuning} \label{sec:unified_prompt}
\begin{figure}[h]
\begin{minipage}{0.74\textwidth} 
\begin{center}
    \includegraphics[width=0.99\textwidth]{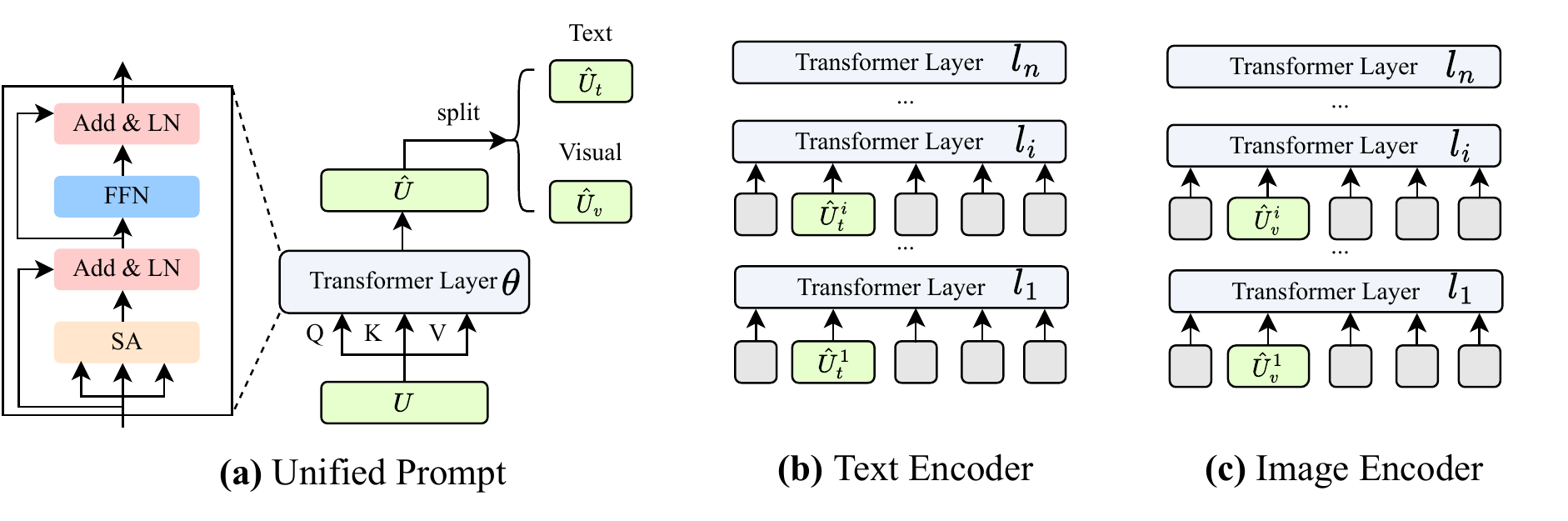}
\end{center}
\end{minipage}\hspace{0.01\textwidth}
\begin{minipage}{0.24\textwidth} 
\centering
\caption{The architecture of (\textbf{a}) our unified prompt $\mU$ that is applied to (\textbf{b}) CLIP \emph{text} encoder and (\textbf{c}) CLIP \emph{image} encoder.}
    \label{fig:architecture}
\end{minipage}
\end{figure}

Driven by our analysis, we devise a simple yet effective multi-modal \textbf{U}nified \textbf{P}rompt \textbf{T}uning (UPT) approach for adapting VL models.
Specifically, instead of introducing two sets of isolated modality-specific prompts (\ie, $\mT$ in Eq.~(\ref{eq:text_prompt}) and $\mV$ in Eq.~(\ref{eq:visual_prompt})) for the text and visual encoders, we consider learning a set of unified modality-agnostic prompts for tuning VL models. As shown in Fig.~\ref{fig:architecture}, we define a set of learnable prompts $\mU \in \mathbb{R}^{d \times n}$ with length $n$. Rather than na\"{i}vely appending the unified prompts into the text and visual encoders, we employ a lightweight Transformer layer $\theta$ to transform unified prompts $\mU$ as follows:
\begin{equation}
    \begin{aligned}
    \mU^{\prime} &= \operatorname{SA}\left( \mU \right) + \mathrm{LN}\left( \mU \right), \\
    \hat{\mU} &= \operatorname{FFN}\left(\mathrm{LN}\left(\mU^{\prime}\right)\right) + \mathrm{LN}\left(\mU^{\prime}\right),
    \end{aligned}
\end{equation}

where the self-attention operator~$\operatorname{SA}$, feed-forward network~$\operatorname{FFN}$ and layer normalization~$\mathrm{LN}$ are applied to obtain the transformed prompts $\hat{\mU}$. The self-attention module in the lightweight Transformer layer allows beneficial interaction between two modalities, so as to maximize the complementary effects. 
Our unified prompts can be introduced into multiple layers of VL models. In particular, for each $i$-th layer of text and image encoders, we consider learning a set of layer-wise prompts $\mU^{i}$, and split transformed $\hat{\mU^{i}}$ into two parts $\hat{\mU^{i}} = \{ \hat{\mU}^{i}_{t}, \hat{\mU}^{i}_{v} \}$, sending into the text and visual encoders respectively. During downstream training, we froze both the text and visual encoder ($\psi$ and $\phi$) and only optimize the unified prompts $\mU$ and the lightweight Transformer layer $\theta$. In this way, both the dynamic classifiers $\mW$ and visual features $\vz$ in Eq.~(\ref{eq:cos}) are effectively tuned for reliable prediction in the downstream task.
\if\mycmd0
As shown in Fig.~\ref{fig:motivation}(d), 
\else
As shown in Fig.~\ref{fig:motivation}~(d) and (h), 
\fi
our unified prompts can simultaneously obtain well-aligned text classifiers and separable visual features compared with single-modal counterparts.
\section{Experiments}
In this section, we conduct experiments under two problem settings, \ie, (i) few-shot image classification (Sec.~\ref{sub:fewshot}) and (ii) domain generalization (Sec.~\ref{sub:crosseval}).
We also present ablation studies in Sec.~\ref{sec:ablation} on several design choices.

\noindent \textbf{Baselines.} 
We compare our approach against the following methods:
(1) \textbf{Zero-shot CLIP}. This baseline uses hand-crafted text prompt templates and does not involve any prompt-learning strategies.
(2) \textbf{Single-modal Prompt Tuning} methods, including CoOp~\citep{zhou2021coop}, for the text modality, and VPT~\citep{jia2022visual}, for the visual modality. In the domain generalization setting, we further compare with CoCoOp~\citep{zhou2022cocoop}, which improves CoOp's generalization performance with an input-conditional design. For VPT, we report the results of both the shallow and deep variants, as described in Sec.~\ref{sub:preliminaries}.

\subsection{Few-shot Learning} \label{sub:fewshot}
In this section, we measure a model's generalization ability by conducting prompt tuning using different strategies, with just a limited amount of labeled examples per-class in the specific downstream task.
Detailed implementation is presented in Appendix~\ref{sec:app_impl}.

\begin{figure*}[t]
    \centering
    \includegraphics[width=.98\textwidth]{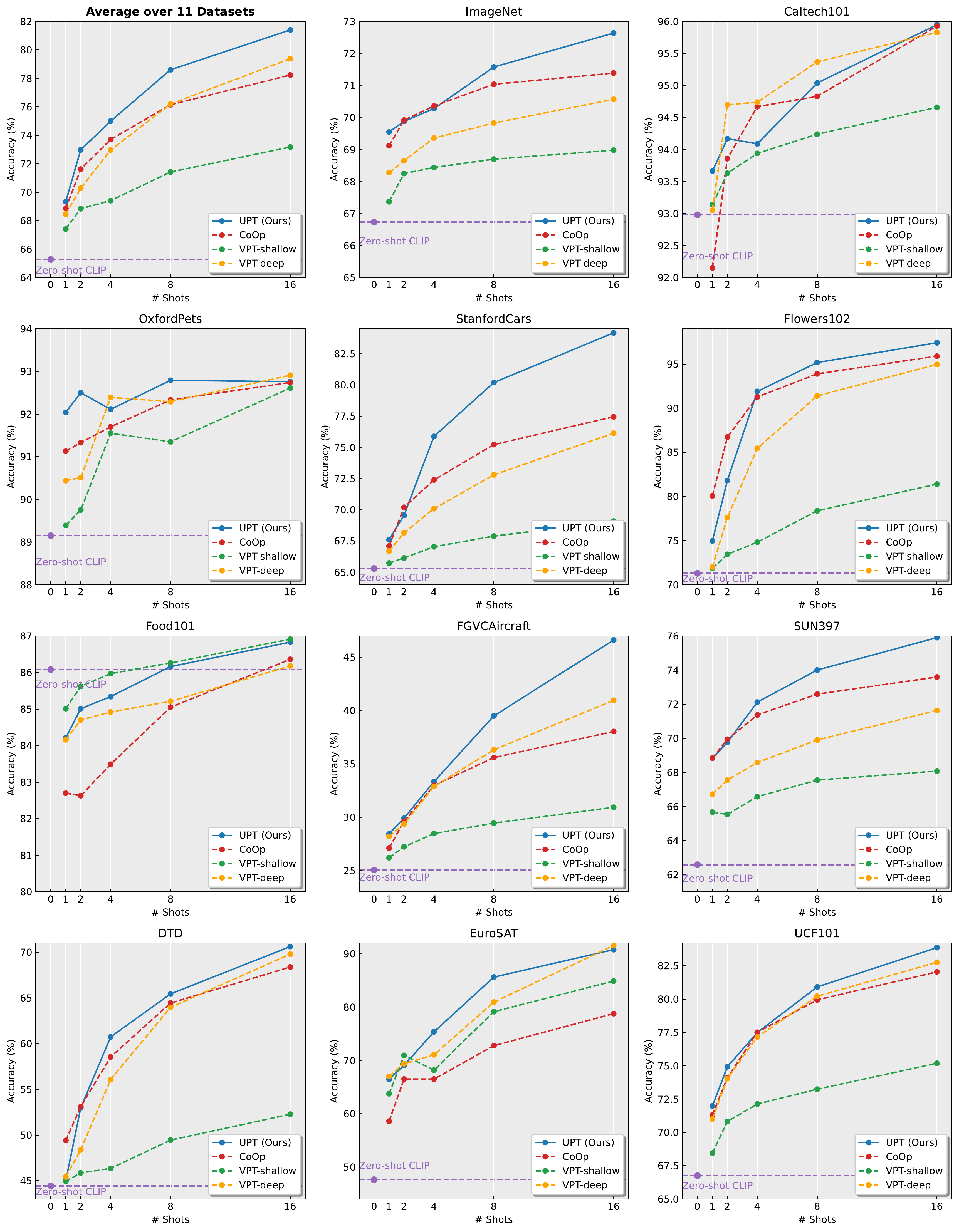}
    \caption{\textbf{Main results over 11 datasets under the few-shot learning setting.}
    We report the average accuracy (\%) of 1/2/4/8/16 shots over three runs.
    Overall, the proposed UPT (\textcolor{blueline}{blue line}) achieves apparent improvements compared with the Zero-shot CLIP and single-modal prompt tuning baselines (CoOp and VPT).}
    \label{fig:main_results}
    \vspace{-12pt}
\end{figure*}

\noindent \textbf{Datasets.}
We follow~\citet{zhou2022cocoop} to use 11 datasets (ImageNet~\citep{deng2009imagenet}, Caltech101~\citep{fei2004learning}, OxfordPets~\citep{parkhi2012cats}, StanfordCars~\citep{krause20133d}, Flowers102~\citep{nilsback2008automated}, Food101~\citep{bossard2014food}, FGVC-Aircraft~\citep{maji2013fine}, SUN397~\citep{xiao2010sun}, UCF101~\citep{soomro2012ucf101}, DTD~\citep{cimpoi2014describing}, EuroSAT~\citep{helber2019eurosat}) as our benchmarks.
Following~\citet{zhou2021coop}, we use the few-shot evaluation protocol selecting 1/2/4/8/16 shots for training and the whole test set for evaluation.
We report averaged results over three runs with different random seeds to reduce the variance.
The detailed results are shown in Fig.~\ref{fig:main_results}.

\noindent \textbf{Limitation of Single-modal Baselines.}
Figure~\ref{fig:main_results} shows that the performance improvements of existing \emph{text} prompt tuning method CoOp and \emph{visual} prompt tuning method VPT are not consistent across different datasets.
In particular, CoOp obtains better performance than VPT on some datasets, such as StanfordCars and SUN397.
However, for other datasets with high intra-class visual variances, VPT is much more effective than CoOp.
For instance, on the EuroSAT dataset, VPT-deep beats CoOp by over 12\%.
The discrepancy of previous single-modal baselines is also consistent with our motivation in Fig.~\ref{fig:comparison}(d) and Fig.~\ref{fig:comparison}(e).
According to VPT~\citep{jia2022visual}, VPT-deep is more effective than VPT-shallow, and our experimental results also verify this point.
We later show that VPT-shallow obtains much stronger performance than the VPT-deep in the domain generalization setting~(Sec.~\ref{sub:crosseval}).

\noindent \textbf{UPT vs. Single-modal Baselines.}
Our UPT achieves clear advantages over the single-modal prompt-tuning counterparts CoOp and VPT, as suggested by the averaged performance (top-left of Fig.~\ref{fig:main_results}).
In general, the average performance gap between UPT and baselines increases with the shot number available for prompt tuning.
Specifically, UPT obtains 0.48/1.36/1.29/2.46/3.19(\%) accuracy improvements compared with the \emph{text} prompt tuning method CoOp on 1/2/4/8/16 shots settings.
Similarly, UPT achieves 0.89/2.70/2.03/2.40/2.01(\%) accuracy gains over the \emph{visual} prompt tuning approach VPT-deep.
Notably, UPT significantly boosts the performance over CoOp and VPT-deep on challenging large datasets, such as ImageNet with 1,000 classes and SUN397 with 397 categories.
UPT also surpasses CoOp and VPT-deep on fine-grained datasets such as StanfordCars and FGVC Aircraft.
We also observe that UPT shows less improvement on the two datasets~(OxfordPets and Food101), possibly caused by the noisy training data~\citep{zhou2021coop,bossard2014food}.
Overall, the experimental results in Fig.~\ref{fig:main_results} demonstrate the effectiveness of our proposed UPT.

\subsection{Domain Generalization} \label{sub:crosseval}
Pre-trained VL models like CLIP have shown strong generalization ability.
However, the prompt tuned on a specific downstream dataset may hinder the generalization ability on categories outside the training set.
In this section, we evaluate the generalization ability of different prompt tuning methods to out-of-distribution~(OOD) data.

\noindent \textbf{Datasets.} We follow~\citep{zhou2021coop} to use five datasets (ImageNet~\citep{deng2009imagenet}, ImageNet V2~\citep{recht2019imagenet}, ImageNet-Sketch~\citep{wang2019learning}, ImageNet-A~\citep{hendrycks2021natural} and ImageNet-R~\citep{hendrycks2021many}) for evaluation.
Following the protocol, we train a model on ImageNet and evaluate it on four other variants of ImageNet with their domains shifted.

\noindent \textbf{Results.}
Table~\ref{tab:ood} summarizes the results.
We report the average accuracy on both the source and target datasets~(penultimate column), and the OOD average accuracy on target datasets~(last column).
The results show that VPT-shallow~(row \rowNumber{\#2}) achieves higher OOD accuracy than VPT-deep~(row \rowNumber{\#3}), and \emph{text} prompt tuning methods outperform \emph{visual} prompt tuning approaches.
Furthermore, the proposed UPT (row \rowNumber{\#5}) is generally a better option than single-modal baselines (rows \rowNumber{\#1-\#4}) and obtains comparable performance with CoCoOp.
Our UPT achieves the best results four times on five target datasets, showing that UPT is a reliable prompt tuning method among its competitors in the domain generalization setting.

\begin{table}[t]
\centering
\caption{\textbf{Main results under the domain generalization setting.} We report the average accuracy (\%) of 16 shots over three runs.
The \textbf{best} and \textbf{second best} methods are highlighted in \colorbox{rred}{red} and \colorbox{oorange}{orange}, respectively.}
\tabstyle{7.5pt}
\begin{tabular}{ll c cccc cc}
\toprule
\multirow{2}{1em}{\rowNumber{\#}} & \multirow{2}{4em}{Method} & Source & \multicolumn{4}{c}{Target} & \multirow{2}{4em}{Average} & \multirow{2}{4em}{\emph{OOD} Average} \\
\cmidrule(lr){3-3} \cmidrule(lr){4-7}
&  & ImageNet & -V2 & -S & -A & -R & ~ \\
\midrule
\rowNumber{1} & CoOp & \cellcolor{oorange} 71.51 & \cellcolor{oorange} 64.20 & 47.99 & 49.71 & 75.21 & 61.72 & 59.28 \\
\rowNumber{2} & CoCoOp & 71.02 & 64.07 & \cellcolor{rred} \textbf{48.75} & \cellcolor{oorange} 50.63 & \cellcolor{oorange} 76.18 & \cellcolor{oorange} 62.13 & \cellcolor{oorange} 59.91 \\
\rowNumber{3} & VPT-shallow & 68.98 & 62.10 & 47.68 & 47.19 & 76.10 & 60.38 & 58.27 \\
\rowNumber{4} & VPT-deep & 70.57 & 63.67 & 47.66 & 43.85 & 74.42 & 60.04 & 57.40 \\
\midrule
\rowNumber{5} & UPT & \cellcolor{rred} \textbf{72.63} & \cellcolor{rred} \textbf{64.35} & \cellcolor{oorange} 48.66 & \cellcolor{rred} \textbf{50.66} & \cellcolor{rred} \textbf{76.24} & \cellcolor{rred} \textbf{62.51} & \cellcolor{rred} \textbf{59.98} \\
\bottomrule
\end{tabular}
\label{tab:ood}
\vspace{-12pt}
\end{table}

\subsection{Ablation Studies} \label{sec:ablation}

\begin{figure*}
\centering
\includegraphics[width=.9\textwidth]{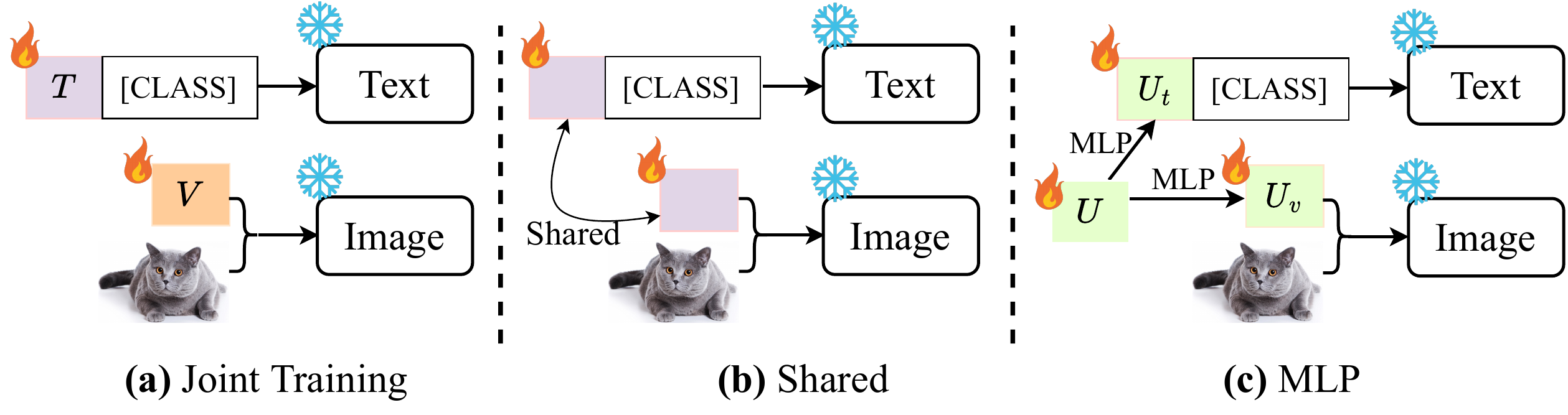}
\caption{Ablation studies on different design choices. \textbf{(a)}: jointly train the existing \emph{text} and \emph{visual} prompt tuning approaches; \textbf{(b)}: shared prompts for all modalities; \textbf{(c)}: using two MLP layers to generate the prompts.}
\label{fig:abla_unified}
\end{figure*}

\begin{table*}[t]
    \tabstyle{3.5pt}
    \caption{Ablation studies on different multi-modal prompt design choices in Fig.~\ref{fig:abla_unified} over 11 datasets.
    We report the accuracy results under the 16 shots setting.
    The \textbf{best} and \textbf{second best} methods are highlighted in \colorbox{rred}{red} and \colorbox{oorange}{orange}, respectively.}
    \label{tab:abla_unified}
    \begin{tabular}{ll ccccccccccccccc}
    \toprule
    \\ 
    \rowNumber{\#} & \rotbox{Method} & \rotbox{ImageNet} & \rotbox{Caltech101} & \rotbox{OxfordPets} & \rotbox{StanfordCars} & \rotbox{Flowers102} & \rotbox{Food101} & \rotbox{FGVCAircraft} & \rotbox{SUN397} & \rotbox{DTD} & \rotbox{EuroSAT} & \rotbox{UCF101} & \rotbox{\emph{Average}} \\
    \midrule
    \rowNumber{1} & CoOp & 71.36 & \cellcolor{oorange}95.93 & 92.74 & 77.45 & \cellcolor{oorange}95.90 & 86.36 & 38.04 & 73.59 & 68.38 & 78.77 & 82.04 & 78.24 \\
    \rowNumber{2} & VPT-shallow & 68.98 & 94.66 & 92.61 & 69.09 & 81.40 & 86.91 & 30.93 & 68.08 & 52.28 & 84.87 & 75.19 & 73.18 \\
    \rowNumber{3} & VPT-deep & 70.57 & 95.83 & 92.91 & 76.13 & 94.96 & 86.18 & \cellcolor{oorange}40.96 & 71.63 & \cellcolor{oorange}69.79 & \cellcolor{rred}\textbf{91.53} & 82.76 & \cellcolor{oorange}79.39 \\
    \midrule
    \rowNumber{4} & Joint Training & 71.42 & 95.84 & \cellcolor{oorange}93.34 & \cellcolor{oorange}79.02 & 95.25 & 86.55 & 40.56 & 74.17 & 67.83 & 78.94 & \cellcolor{oorange}82.81 & 78.70 \\
    \rowNumber{5} & Shared & \cellcolor{oorange}71.46 & 95.50 & 92.99 & 78.66 & 95.55 & \cellcolor{oorange}86.67 & 39.18 & \cellcolor{oorange}73.64 & 67.69 & 73.36 & 82.06 & 77.88 \\
    \rowNumber{6} & MLP & 71.00 & 95.59 & \cellcolor{rred}\textbf{93.74} & 75.88 & 93.38 & \cellcolor{rred}\textbf{87.20} & 37.17 & 72.74 & 67.31 & \cellcolor{oorange}90.66 & 81.43 & 78.73 \\
    \midrule
    \rowNumber{7} & UPT (Ours) & \cellcolor{rred}\textbf{72.63} & \cellcolor{rred}\textbf{95.94} & 92.95 & \cellcolor{rred}\textbf{84.33} & \cellcolor{rred}\textbf{97.11} & 85.00 & \cellcolor{rred}\textbf{46.80} & \cellcolor{rred}\textbf{75.92} & \cellcolor{rred}\textbf{70.65} & 90.51 & \cellcolor{rred}\textbf{84.03} & \cellcolor{rred}\textbf{81.44} \\
    \bottomrule
    \end{tabular}
\vspace{-12pt}
\end{table*}

\noindent \textbf{Comparison with the Joint Training Baseline.}
As shown in Fig.~\ref{fig:abla_unified}(a), a straightforward approach for multi-modal prompts is tune the \emph{text} prompt (using CoOp) and \emph{visual} prompt~(using VPT) jointly.
We investigate the effectiveness of such joint training scheme, and report its results in Table~\ref{tab:abla_unified} row \rowNumber{\#4}.
From the results, we see that such a joint training solution performs slightly worse than the \emph{visual} prompt tuning method VPT-deep (78.70\% vs 79.39\%), and shows inferior performance to our UPT.

\noindent \textbf{Shared Prompts for \emph{Text} and \emph{Visual} Modalities.}
We also investigate the results of directly sharing prompts for different modalities.
As shown in Fig.~\ref{fig:abla_unified}(b), the shared prompts will be optimized for both \emph{text} and \emph{visual} modalities. This scheme differs from the proposed UPT, where the shared prompts are transformed with self attention.
Experimental results are presented in Table~\ref{tab:abla_unified} row \rowNumber{\#5}, and we observe that such a prompt sharing strategy achieves worst performance among all the methods.

\noindent \textbf{MLP Baseline.}
For our proposed UPT, we use a Transformer layer with the self-attention operator to partially share the hyper-parameters for different modalities.
Here, we study a simpler design that generates the unified prompts with two MLP layers.
Results are presented on Table~\ref{tab:abla_unified} row \rowNumber{\#6}. The MLP baseline is still competitive, yielding best performance on two datasets. Nonetheless, the average results on 11 datasets is still poorer than the proposed self-attention based approach.

\subsection{Qualitative Results}

\begin{figure*}[t]
    \centering
    \includegraphics[width=.99\textwidth]{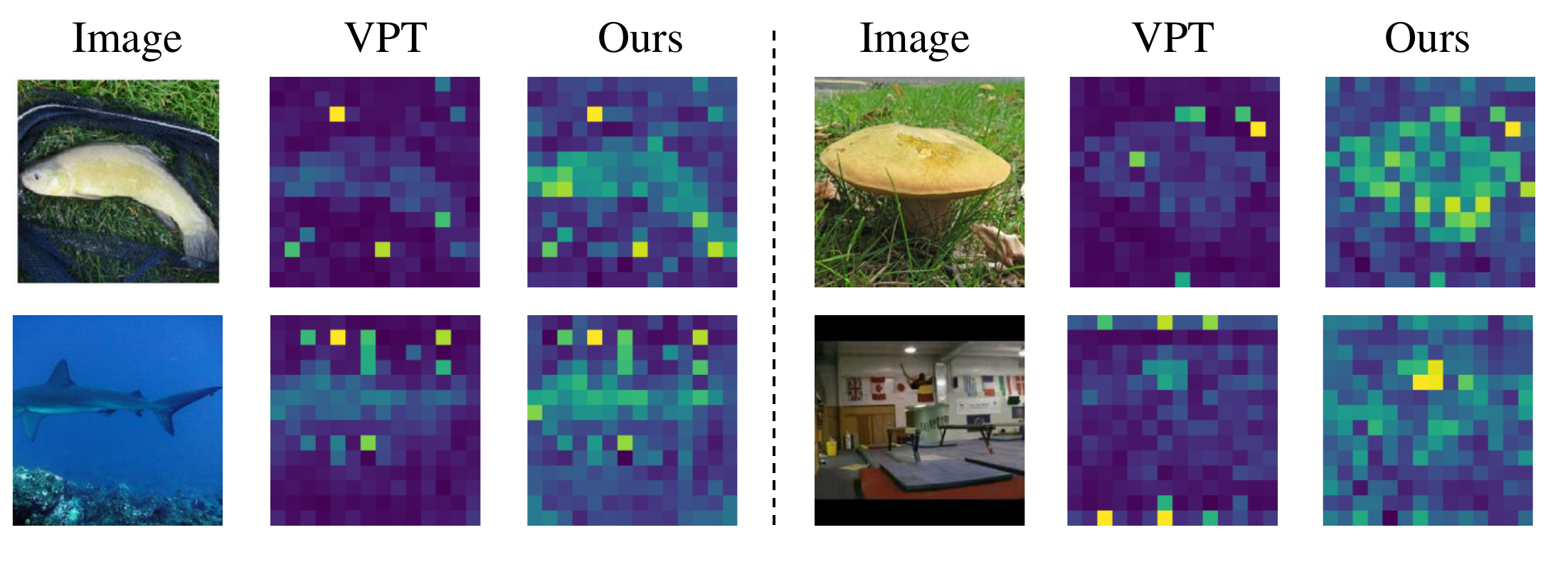}
    \caption{Visualization of attention response map between visual prompts and image patch tokens. The images are test images from ImageNet. We visualize the self-attention module from the last block of ViT of the CLIP image encoder.
    }
    \label{fig:vis_attn}
    \vspace{-12pt}
\end{figure*}

While it is hard to visualize what have been learned during text prompt tuning, it is possible to visualize the visual prompts learned by VPT and UPT following the self-supervised learning method, DINO~\citep{caron2021emerging}. In particular, for each layer of the Vision Transformer~(ViT), we can compute the self-attention response map of visual prompts and image patch tokens.
Figure~\ref{fig:vis_attn} compares such response maps by VPT and the proposed UPT.
We find that UPT shows stronger self-attention responses compared with VPT.
This could be the possible reason why UPT achieves better performance on the few-shot learning and the OOD generalization settings.

\section{Related Work}

\noindent \textbf{Vision-Language Models.}
Recent vision-language pre-trained models~\citep{radford2021learning,jia2021scaling} use the contrastive loss to align an image encoder (\eg, ViT~\citep{dosovitskiy2020vit}) and a text encoder (\eg, BERT~\citep{kenton2019bert}) in a common feature space. These vision-language models are trained on web-scale image-text pairs and are transferable across various downstream tasks such as point cloud classification~\citep{zhang2022pointclip}, video classification~\citep{qian2022multimodal}, object detection~\citep{gu2021open,du2022learning,zhou2022detecting,zang2022open} and semantic segmentation~\citep{ghiasi2021open}. In this work, we aim to explore how to adapt the CLIP model to the downstream few-shot recognition task.

\noindent \textbf{Text Prompt Tuning.}
The concept of prompt tuning was first proposed in the NLP area~\citep{liu2021pre,gao2021making,li2021prefix,lester2021power}. In particular, a text prompt refers to a task-specific template for language models. For example, in sentiment analysis, the template might be ``\texttt{I [MASK] the movie.}'' where the mask placeholder will be filled with either ``love'' or ``hate.'' Common practices in text prompt tuning include (i) searching for a specific word in the dictionary, known as \textit{hard} prompt learning~\citep{gao2021making}, or (ii) turning masked tokens into learnable vectors, known as \textit{soft} prompt learning~\citep{li2021prefix,lester2021power}. Text prompt tuning has also been applied in computer vision after the emergence of large vision-language models (\eg, CLIP~\citep{radford2021learning}), which are too big to fine-tune. A representative work is CoOp~\citep{zhou2021coop}, which turned the input context tokens in CLIP's text branch into learnable vectors for adapting CLIP to downstream image recognition. Other follow-ups of CoOp include CoCoOp~\citep{zhou2022cocoop}, DualCoOp~\citep{sun2022dualcoop}, ProGrad~\citep{xing2022class}, and ProDA~\citep{lu2022prompt}.

\noindent \textbf{Visual Prompt Tuning.}
The idea of visual prompt tuning is to adapt large pre-trained Vision Transformers~\citep{dosovitskiy2020vit} by adding learnable parameters in the visual input space, which is analogous to text prompt tuning in NLP. VPT~\citep{jia2022visual} and Visual Prompting~\citep{bahng2022visual} both add trainable tokens to the input of Transformer models. A recent work, NOAH~\citep{zhang2022neural}, uses neural architecture search algorithms to identify the optimal configuration of prompt modules. In comparison to the unimodal prompt learning methods discussed above, our paper provides a timely study on how to achieve a better trade-off using multimodal prompt learning.
\section{Conclusion}
With the rapid scaling of vision models along the size dimension, efficient downstream adaptation methods have become essential for facilitating large-scale deployment of vision models in the wild. Our paper provides a timely and comprehensive study on how to adapt large vision-language models like CLIP from the \textit{prompt learning} perspective. In particular, our study unveils that the previous unimodal prompt tuning methods do not work consistently well across a wide range of vision datasets. In contrast, the proposed UPT method, despite having a simple design, achieves a better trade-off compared with the unimodal counterparts. The results suggest that one should exploit correspondences between different modalities for prompt learning.

On the other hand, the results achieved by UPT are by no means perfect: in the ablation studies we observe that some alternative designs, such as using MLP instead of Transformer, might sometimes give better performance. Overall, we believe multimodal prompt learning is a promising framework to build upon, and we expect more improvements can be achieved with more advanced (and efficient) designs.

\bibliography{reference}
\bibliographystyle{iclr2023_conference}

\newpage
\appendix

\noindent {\large \textbf{Appendix}}

\section{Intra-/Inter- Class Variance}
\label{sec:app_var}
In this section, we provide the implementation details about how we compute the \textit{intra-class} visual variance and \textit{inter-class} text variance for different datasets (Fig.1~(d)(e) in the main paper.

\noindent \textbf{Intra-class Visual Variance}.
Given one dataset with $k$ classes in total, for each image $\vx$ that belongs to class $c$, we first use the CLIP image encoder $\phi$ to extract the corresponding image feature $f_{\phi}(\vx)$. 
Then we get the intra-class variance of class $c$ as:
\begin{equation}
    \text{var}_{c} = \frac{1}{|| X_{c} ||} \sum_{x \in X_c} \left( f_{\phi}(\vx) - \bar f_{\phi}(\vx) \right)^{2},
\end{equation}
where $X_c$ denotes to the set of images that have the ground-truth class label $c$, and $\bar f_{\phi}(\vx)$ refers to the mean values of class $c$. Then we can compute the intra-class variance $\text{Var}_{\text{v}}$ for all the $k$ classes as
\begin{equation}
   \text{Var}_{\text{v}} =  \frac{1}{k} \sum_{c=1}^{k} \text{var}_{c}.
\end{equation}

\noindent \textbf{Inter-class Text Variance}.
For each dataset, we first compute the CLIP text features $\vw$ of classs $c$, and the mean value $\bar \vw$ of all the $k$ classes.
Then we get the inter-class text variance $\text{Var}_{\text{t}}$ as:
\begin{equation}
   \text{Var}_{\text{t}} = \frac{1}{k} \sum_{c=1}^{k}(w_{c} - \bar w)^{2}.
\end{equation}

\section{Implementation Details} \label{sec:app_impl}
Our implementation is based on the source code of CoOp~\citep{zhou2021coop}.
We use ViT-B/16 as the CLIP backbone~\citep{radford2021learning}.
Following~\citet{zhou2022cocoop}, we set the context length of CoOp as $m=4$ (same for VPT).
For Zero-shot CLIP and VPT, we use the default prompt template, ``\texttt{a photo of a {[CLS]}.}''
We use SGD as the optimizer, with an initial learning rate of 0.002, which is decayed by the cosine annealing rule.
The batch size is set to 32 for all datasets.

\end{document}